\documentclass[11pt,a4paper]{article}
\usepackage[hyperref]{emnlp-ijcnlp-2019}
\usepackage{times}
\usepackage{latexsym}

\usepackage{url}

\usepackage{booktabs}
\usepackage[plain]{fancyref}
\usepackage[utf8]{inputenc}
\usepackage{amsmath}
\usepackage{algorithmicx}
\usepackage{algorithm}
\usepackage[noend]{algpseudocode}

\usepackage[disable]{todonotes}   

\aclfinalcopy 

\title{Transductive Auxiliary Task Self-Training for Neural Multi-Task Models}

\newcommand*{\affaddr}[1]{#1} 
\newcommand*{\affmark}[1][*]{\textsuperscript{#1}}

\author{\parbox{\linewidth}{\centering
Johannes Bjerva{\rm\affmark[1]}, Katharina Kann{\rm\affmark[2]}, 
Isabelle Augenstein{\rm\affmark[1]}} \vspace{.12cm}
\\
\affaddr{\affmark[1]Department of Computer Science, University of Copenhagen, Denmark
\\
\affaddr{\affmark[2]Center for Data Science, New York University, USA}} \vspace{.1cm}
\\
\affaddr{\texttt{bjerva,augenstein@di.ku.dk, kann@nyu.edu}}}

\date{}

\begin{document}
\maketitle

\begin{abstract}

Multi-task learning and self-training are two common ways to improve a machine learning model's performance
in settings with limited training data. Drawing heavily on ideas from those two approaches, 
we suggest transductive auxiliary task self-training: training a multi-task model on (i) a combination of main and auxiliary task training data, and (ii) test instances with auxiliary task labels which a single-task version of the model has previously generated. 
We perform extensive experiments 
on 86 combinations of languages and tasks.
Our results are that, on average, transductive auxiliary task self-training improves absolute accuracy by up to $9.56\%$ over the pure multi-task model for dependency relation tagging and by up to $13.03\%$ for semantic tagging.
\end{abstract}

\begin{figure*}[htbp]
  \centering
  \includegraphics[width=.9\textwidth]{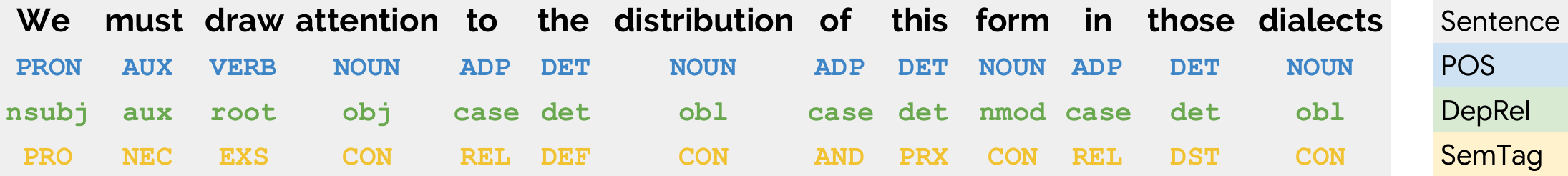}
  \caption{\label{fig:ex1}POS tags, DepRel labels and semantic tags for an example sentence.}
\end{figure*}  

\section{Introduction}
\label{intro}
When data for certain tasks or languages is not readily available, different approaches exist to leverage other resources
for the training of machine learning models.
Those are commonly either instances from a related task or unlabelled data:
During \textbf{multi-task training} \cite{Caruana:93}, 
a model learns from examples of multiple related tasks at the same time and can therefore benefit from a larger overall number of training instances.
\textbf{Self-training} \cite{Yar95,Riloff+Wiebe+Wilson:03a}, in contrast, denotes the process of iteratively training a model, using it to label new examples, and adding the most confident ones to the training set before repeating the training.
As data without gold standard annotations is used, self-training can be considered a special case of semi-supervised training.

In this work, we propose \textbf{transductive auxiliary task self-training}, based on a combination of multi-task training and self-training:
We use the available auxiliary task data to obtain a high-performing single-task model for the auxiliary task,
which we then use to label the main task test set with auxiliary task labels.
Subsequently, we train a multi-task model on both tasks, while including instances with the newly generated silver
standard auxiliary task labels.

Transductive auxiliary task self-training is an extremely cheap procedure, requiring only small amounts of additional computing time, compared to the obvious alternative of manually producing more labels. 
Our approach is \textbf{transductive} since the model generalises from specific training examples to specific test examples.
In particular, training on auxiliary task labels for the test set, which have been produced by the single-task model, yields a final multi-task model, which satisfies the defining criterion of transductive inference that predictions depend on the test data \cite{vapnik98theory}. Note that we do not require
gold standard test labels for either task.




In addition to presenting our method, we investigate three research questions (\textbf{RQ}s):

{\bf RQ 1:} For which tasks and dataset sizes 
does transductive auxiliary task self-training 
help most?

{\bf RQ 2:} Can a model trained with our cost-free transductive auxiliary task self-training perform similarly to or better than a model trained on additional manual annotations for the auxiliary task?

{\bf RQ 3:} Even without considering reduced costs, are there scenarios where it is better to perform transductive auxiliary task self-training than adding more main task examples?

In order to find generalisable answers to these research questions, we experiment with several tasks, languages and numbers of training samples. 
We consider the low-level auxiliary task of part-of-speech tagging and two main tasks: dependency relation (DepRel) tagging and semantic tagging.
We furthermore compare with an unsupervised auxiliary task baseline, to show that our results are not simply a result of domain adaptation effects.
We experiment on 41 languages, yielding a total of 86 unique language--task combinations.
We find that, on average, transductive auxiliary task self-training improves absolute accuracy by up to $9.56\%$ and $13.03\%$ over the pure multi-task model for DepRel tagging and semantic tagging, respectively.
\section{Neural Sequence Labelling}
\subsection{Tasks}
Figure~\ref{fig:ex1} shows a sentence
with annotations for the three linguistic tasks considered in this paper, which we will describe in the following.\footnote{The example is taken from PMB document 01/3421, which has gold standard SemTags. The UD POS and DepRel tags were obtained using UD-Pipe \citep{udpipe}.}

\textbf{Part-of-speech (POS) tagging} is the task of assigning morpho-syntactic tags to each word in a sentence. 
We use it as an auxiliary task, 
since respective datasets are available for many languages.
It is also a relatively easy task, with state-of-the-art models typically achieving over $95\%$ accuracy \citep{plank:2016}.
We use the Universal Dependencies (UD) POS tag set \citep{nivre:2016}.

\textbf{Dependency relation (DepRel) labelling} 
is the task of assigning dependency labels to each word in a sentence.
In our experiments, we use the Universal Dependencies labels \citep{nivre:2016}.
We use this task as a main task.
Both this task and POS tagging are morpho-syntactic tasks.

\textbf{Semantic Tagging (SemTag)} is the task of assigning a semantic tag to each word in a sentence.
We use the labels from the Parallel Meaning Bank (PMB, \citet{pmb,bjerva:2016:semantic}).
This tag set was designed for multilingual semantic parsing and, therefore, to generalise across languages.
As this task is relatively challenging, we use it as a main task.
While the UD data is available for 41 languages, the PMB data is only available for four (English, Italian, Dutch, and German).

\textbf{FreqBin} is the task of predicting the binned frequency of a word, as introduced by \citet{plank:2016}. We use this task as an unsupervised auxiliary baseline, with frequencies obtained from our training data. 

\subsection{Model Architecture}
We approach sequence labelling by using a variant of a bidirectional recurrent neural network, which uses both preceding and succeeding context when predicting the label of a word.
This choice was made as such models at the same time obtain high performance on all three tasks and lend themselves nicely to multi-task training
via hard parameter sharing. 
This system is based on the hierarchical bi-LSTM of \citet{plank:2016} and is implemented using DyNet \citep{dynet}.
On the subword-level, the LSTM is bi-directional and operates on characters \cite{ballestros:2015,ling:2015}.
Second, a context bi-LSTM operates on the word level, from which output is passed on to a  classification layer.



Multi-task training is approached using hard parameter sharing \citep{Caruana:93}.
We consider $T$ datasets, each containing pairs of input-output sequences $(w_{1:n}, y^t_{1:n})$, $w_i \in V$, $y^t_i \in L^t$.
The input vocabulary $V$ is shared across tasks, but the outputs (tagsets) $L^t$ are task dependent.
At each step in the training process we choose a random task $t$, followed by a randomly chosen batch of training instance.
Each task is associated with an independent classification function, but all tasks share the hidden layers. 
We train using the Adam optimisation algorithm \citep{adam} over a maximum of 10 epochs together with early stopping. 

\section{Transductive Auxiliary Task Self-Training}
Manual annotation of data for main or auxiliary tasks is time-consuming and expensive. Instead,
we propose to use a preliminary single-task model to label the main task test data with auxiliary task labels
which can then be leveraged to train an improved multi-task model.

Transductive auxiliary task self-training is based on two main ideas. 
First, we assume that the auxiliary task is easier than the main task, such that a high performance can be achieved on it.
Hence, the model will be confident about the auxiliary task labels, as is required for self-training.
Second, we choose a transductive approach, because we assume that not all auxiliary task examples will lead to equal improvements on the main task. 
In particular, we expect auxiliary task labels for the test instances to be most useful, 
since information about those instances is most relevant for the prediction of the main task labels on this data.
Similarly to contextualised word representations, this offers an additional signal for the test set instances, as we obtain this through predicted auxiliary labels rather than direct encoding of the context \citep{bert,elmo}.




\subsection{Algorithm}
Our proposed algorithm is shown in Algorithm~\ref{algo:st}.
We start by first training a single-task model on the available auxiliary task training data, which
then predicts labels for the raw input sentences from the main task test set.
Note that we neither observe nor require any labels for this test set, neither for the auxiliary nor for the main task.
The labels which the preliminary single-task model predicts are then added to the train set of the auxiliary task
for training of the final multi-task model.

Although a transductive approach requires training a new model for each test set, sequence-labelling models such as bi-LSTMs
are usually quick to train even on single CPUs, with a full self-training iteration in this paper completing in a matter of hours. 

\begin{algorithm}
\caption{Transductive auxiliary task self-training}\label{algo:st}
\begin{algorithmic}[1]
\State $train_{aux} \gets$ aux. task train data
\State $train_{main} \gets$ main task train data
\State $testinp_{main} \gets$ main task test input
\State $model_{aux} \gets \textbf{train}(train_{aux})$
\For {$sentence \in testinp_{main}$}
\State $l \gets \textbf{label}(sentence, model_{aux})$
\State $train_{aux} = train_{aux} + l$
\EndFor
\State $model_{mtl} \gets \textbf{train}_{mtl}(train_{aux}, train_{main})$
\end{algorithmic}
\end{algorithm}

\section{Experiments}
The experiments described in this section aim at answering the  
research questions raised in \S\ref{intro}, concerned with 
the best settings for transductive auxiliary task self-training, as well as the theoretical question how 
it compares to adding additional (expensive)
gold-standard annotations for the main and the auxiliary tasks.
To ensure that our findings are generalisable, 
we use a large sample of 56 treebanks, covering 41 languages and several domains.
Although this experimental set-up would allow us to run multilingual experiments, we only train monolingual models, and aggregate results across languages and treebanks. 
We investigate three tasks; two of them being morpho-syntactic (POS tagging and DepRel tagging) and one being semantic (semantic tagging).
In all cases, POS is the auxiliary task, and either POS tagging or DepRel tagging is the main task.
Experiments are run in several low-resource settings, varying the amount of main task data.

We run experiments under four conditions, in addition to using an MTL baseline.
We compare (i) adding gold standard test annotations for the auxiliary task only (\textbf{Aux-ST ceiling}), 
(ii) transductive auxiliary task self-training, as described in Algorithm 1 (\textbf{Aux-ST}),
(iii) adding gold standard train annotations for the auxiliary task only (\textbf{Extra Aux}),
or (iv) adding gold standard train annotations for the main task only (\textbf{Extra Main}).
We expect (iii) and (iv) to constitute challenging conditions to beat, as we are in effect giving our model more annotated data, which is normally expensive to come by.

\subsection{Data}
We run experiments on the task-combinations DepRel--POS and Semtag--POS for all available languages and datasets.
Additionally, we reduce our training sets to 10k, 1k, 0.5k, and 0.1k sentences in order to investigate
various low-resource scenarios.
For semantic tagging, the 10k setting is omitted as we do not have enough training data.

 

\setlength{\tabcolsep}{3pt}
\begin{table} 
  \resizebox{\columnwidth}{!}{
  \begin{tabular}{rrr|rrr}
    \toprule
    \textbf{N Main}  & \textbf{MTL} & \textbf{Aux-ST ceiling}  & \textbf{Aux-ST} & \textbf{Extra Aux} & \textbf{Extra Main} \\
    \midrule
    10k  & 86.39 &  *3.79\% & *\textbf{1.97\%} & -0.30\%           & 0.19\%  \\
    1k   & 79.19 &  *7.42\% & *5.39\%          &  *1.74\%          & *\textbf{7.01\%}  \\
    0.5k & 75.77 &  *8.69\% & *6.85\%          &  *2.64\%          & *\textbf{10.17\%} \\
    0.1k & 66.31 & *11.32\% & *9.56\%          &  *4.97\%          & *\textbf{18.15\%} \\
    \midrule
    1k   & 67.82   & n/a & *\textbf{1.74\%}  & 0.05\%     & 0.64\% \\
    0.5k & 63.32   & n/a & *\textbf{4.60\%}  & 0.83\%     & *2.31\%  \\
    0.1k & 50.44   & n/a & *\textbf{13.03\%} & *4.93\%    & *11.58\% \\
    
    \bottomrule
  \end{tabular}
  }
  \caption{\label{tab:results}Macro-averaged changes in accuracy from the MTL baseline for DepRel -- POS (top), SemTag -- POS (bottom). We compare adding gold standard test annotations for the aux task (\textbf{Aux-ST ceiling}), transductive aux task self-training (\textbf{Aux-ST}), adding gold standard train annotations for the aux task (\textbf{Extra Aux}), or for the main task (\textbf{Extra Main}) randomly. Significant ($p<0.05$) differences from the baseline are marked with *.}
\end{table}



\begin{figure}[htbp]
	\includegraphics[width=\columnwidth]{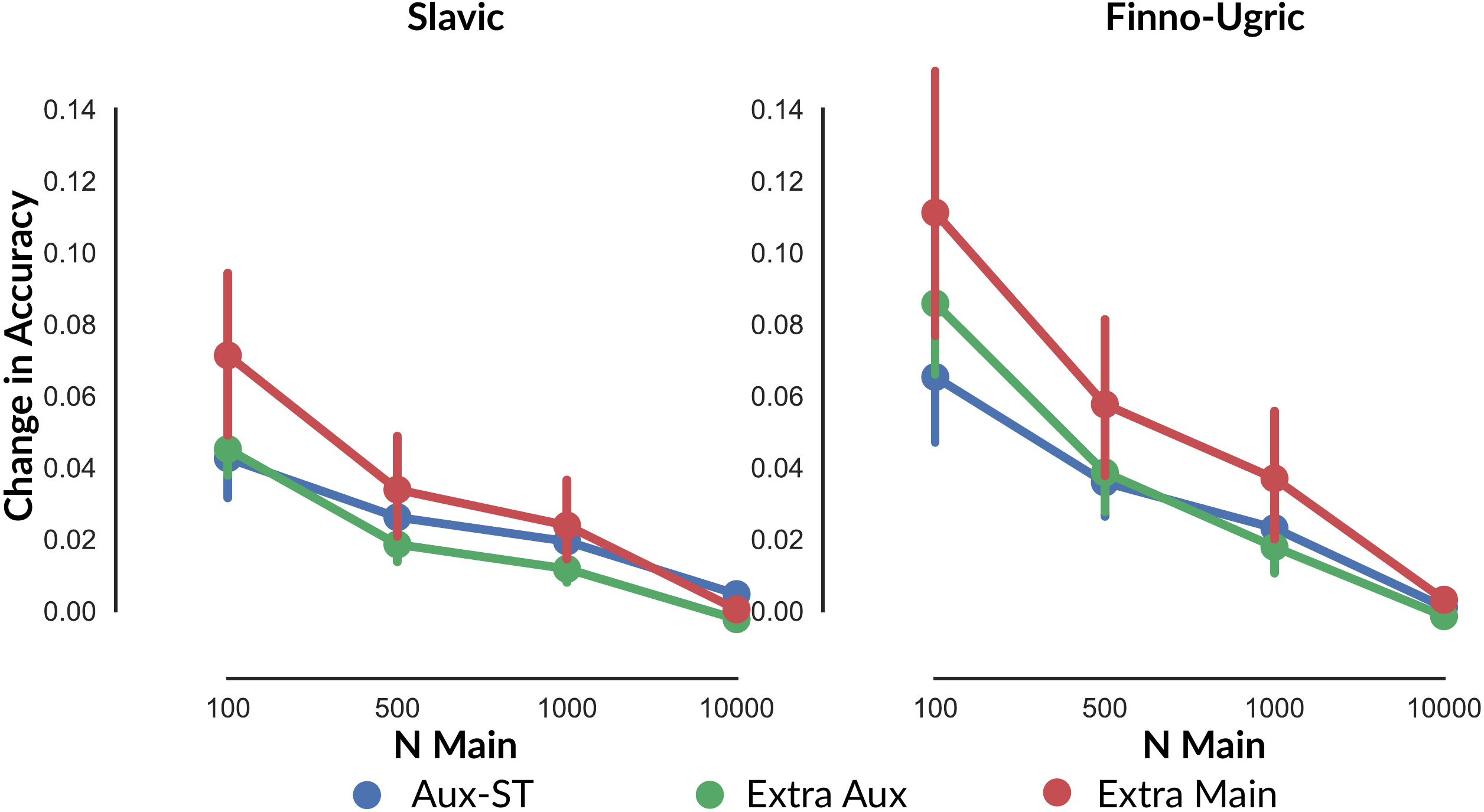}
    \caption{\label{fig:languages}Results for UD treebanks.} 
\end{figure}

\subsection{Results and Discussion}
Table~\ref{tab:results} contains results of the experiments macro-averaged across all languages and treebanks in the UD and across all languages in the PMB. 
Figure~\ref{fig:languages} contains results for two typologically distinct language families, Slavic and Finno-Ugric.

Across all data sizes, self-training on the auxiliary task is significantly better than the baseline multi-task model without self-training.
The results on DepRel tagging show that, when the main task data is sufficiently large, it is more beneficial to do transductive auxiliary task self-training than it is to further increase the size of the main dataset.
For semantic tagging, we find this to hold for all of our training data size settings. 
Our comparison with the FreqBin task does not yield substantial improvements, with mean differences compared to standard MTL at -0.001\% (stdev. 0.022).

To rule out that any gains in the self-training conditions are not due to increased vocabulary, we ran experiments with pre-trained word embeddings which included the raw text from the test set and found no significant differences.
This can be explained by the fact that, although out-of-vocabulary rate is reduced to zero in this condition, the test set is still relatively small.
Thus, the word embeddings do not have much distributional information with which to arrive at good word representations for previously out-of-vocabulary words.


In \textbf{RQ1}, we asked for which task and dataset sizes transductive auxiliary task self-training is most beneficial.
We found benefits across the board, with larger effects when the main task training set is small. 

In \textbf{RQ2}, we asked whether using transductive auxiliary task self-training might even be better than the costly process of manually expanding the data with gold standard auxiliary data for random samples.
We found that this depends on the main task and the size of its training set.
For DepRels, with a low amount of main task data, the largest increase in accuracy is found by adding more main task data.
However, given sufficient main task data, 
 adding highly relevant auxiliary task samples, even ones which are potentially erroneous, is more beneficial.
In the case of semantic tagging, however, transductive auxiliary task self-training is always more beneficial.
As expected, the usefulness of self-training as well as adding extra auxiliary or main task data decreases with increasing dataset size.

In \textbf{RQ3}, we asked whether there are cases in which using auxiliary task data is preferable to annotating and adding more main task samples.
We found that this is the case when using our proposed method of transductive auxiliary task self-training for all training set sizes for semantic tagging, and in the 10k setting for DepRel tagging.

\section{Related Work}


\textbf{Self-training} 
has been shown to be a successful learning approach \cite{NigamGhani:00}, 
e.g., for word sense disambiguation \cite{Yar95} or AMR parsing \cite{conf/acl/KonstasIYCZ17}. 
Samples in self-training are typically selected according to confidence \cite{zhu05survey} which requires a proxy to measure it. This can be the confidence of the model \cite{Yar95,Riloff+Wiebe+Wilson:03a} or the agreement of different models, as used in tri-training \cite{journals/tkde/ZhouL05}. Another 
option is curriculum learning, where selection is based on learning difficulty, increasing the difficulty during learning \cite{Bengio2009}.
In contrast, we build upon the assumption that the auxiliary task examples are ones a model can be certain about.

In \textbf{multi-task learning}, most research focuses on understanding which auxiliary tasks to select, or on how to share between tasks \cite{sogaard2016deep,lin2019choosing,conf/emnlp/RuderP17,augenstein-etal-2018-multi,ruder122019latent}. 
For instance, \citet{conf/emnlp/RuderP17} find that similarity as well as diversity measures applied to the main vs. auxiliary task datasets as a whole are useful in selecting auxiliary tasks.
In the context of sequence labelling, many combinations of tasks have been explored \cite{sogaard-goldberg:2016:P16-2,articleAlonso,bjerva:2017:phd}. 
\citet{ruder122019latent} present a flexible architecture, which learns which parameters to share between a main and an auxiliary task.
One of the few examples where multi-task learning is combined with other methods is the semi-supervised approach 
by \citet{conf/iconip/ChaoS12}, where main task labels are assigned to unlabelled instances which are then added to the main task dataset.
However, to the best of our knowledge, no one has applied self-training to label additional instances with auxiliary task labels.


\section{Conclusion}
We introduced transductive auxiliary task self-training, a straightforward way to improve the performance of multi-task models.
Concretely, we applied the idea of self-training to auxiliary tasks, in order to automatically label the main task
test data with auxiliary task labels which we subsequently included into the training set for multi-task learning. 
In experiments on 41 different languages 
we obtained improvements of up to $9.56\%$ absolute accuracy over the pure multi-task model for DepRel tagging and up to $13.03\%$ absolute accuracy for semantic tagging. 
We further showed that transductive auxiliary task self-training is more effective than randomly choosing additional gold standard auxiliary task data. In some settings, in addition to not needing additional annotation, it even led to a better performing model than adding a comparable amount of extra gold standard main task data.
\bibliography{emnlp-ijcnlp-2019}
\bibliographystyle{acl_natbib}

\end{document}